\documentclass[lettersize,journal]{IEEEtran}
\usepackage{amsmath,amsfonts}
\usepackage[ruled,linesnumbered]{algorithm2e}
\usepackage{array}
\usepackage{textcomp}
\usepackage{stfloats}
\usepackage{url}
\usepackage{multirow}
\usepackage{verbatim}
\usepackage{graphicx}
\usepackage{epstopdf}
\usepackage{cite}
\usepackage{booktabs}
\usepackage{adjustbox}
\usepackage{multirow}
\usepackage{hyperref}
\hypersetup{hidelinks}
\usepackage{mathtools}
\usepackage{adjustbox}
\usepackage{multirow}
\usepackage{enumitem}
\usepackage{multirow}
\usepackage{diagbox}
\usepackage{bm}
\usepackage{amsmath}  
\usepackage{amssymb}  
\usepackage{xcolor}
\usepackage{url}
\usepackage{pifont}
\usepackage{hyperref}
\usepackage{cleveref}
\usepackage{titlesec}
\titleformat{\subsubsection}[runin] {\normalfont\bfseries}{}{0em}{}[:]
\begin{document}

\title{Transferable Dual-Domain Feature Importance Attack against AI-Generated Image Detector}

\author{Weiheng Zhu, Gang Cao,~\IEEEmembership{Member,~IEEE,} Jing Liu, Lifang Yu, and Shaowei Weng~\IEEEmembership{Member,~IEEE}
\thanks{
Manuscript received xxxxxxxxxxx; revised xxxxxxxxxxx; accepted xxxxxxxxxxx. Date of publication xxxxxxxxxxx; date of current version xxxxxxxxxxx. This work was supported in part by the National Natural Science Foundation of China under Grant 62071434, Grant 62262062, and in part by the Fundamental Research Funds for the Central Universities under Grant CUC25GT07. The associate editor coordinating the review of this manuscript and approving it for publication was xxxxxxxxxxxxxxxxx. (Corresponding author: Gang Cao.)
	
Weiheng Zhu and Gang Cao are with the School of Computer and Cyber Sciences, Communication University of China, Beijing 100024, China, and also with the State Key Laboratory of Media Convergence and Communication, Communication University of China, Beijing 100024, China (e-mail: zhuweiheng@mails.cuc.edu.cn,  gangcao@cuc.edu.cn). 

Jing Liu is with the School of Computer Science and Engineering, Hunan University of Information Technology, Changsha 410151, China (e-mail: jingliu1123@163.com).

Lifang Yu is with the Department of Information Engineering, Beijing Institute of Graphic Communication, Beijing 100026, China (e-mail: yulifang@bigc.edu.cn).

Shaowei Weng is with the Fujian Provincial Key Laboratory of Big Data Mining and Applications, Fujian University of Technology, Fuzhou 350118, China (e-mail: wswweiwei@126.com).

Code will be available at https://github.com/multimediaFor/DuFIA .
}
}

\maketitle

\begin{abstract}
	Recent AI-generated image (AIGI) detectors achieve impressive accuracy under clean condition. In view of anti-forensics, it is significant to develop advanced adversarial attacks for evaluating the security of such detectors, which remains unexplored sufficiently. This letter proposes a Dual-domain Feature Importance Attack (DuFIA) scheme to invalidate AIGI detectors to some extent. Forensically important features are captured by the spatially interpolated gradient and frequency-aware perturbation. The adversarial transferability is enhanced by jointly modeling spatial and frequency-domain feature importances, which are fused to guide the optimization-based adversarial example generation. Extensive experiments across various AIGI detectors verify the cross-model transferability, transparency and robustness of DuFIA.
\end{abstract}


\IEEEpeerreviewmaketitle

 \section{Introduction}
With rapid progress of generative models like generative adversarial network (GAN)~\cite{goodfellow2014generative} and diffusion model (DM)~\cite{ho2020denoising}, the AI-generated image (AIGI) exhibits photo-realistic visual quality, making it indistinguishable from real images to the human eye. In response, many AIGI detection algorithms~\cite{gragnaniello2021gan,ojha2023towards,cozzolino2024raising,chen2024drct,corvi2023detection,rajan2024aligned,koutlis2024leveraging} have been developed to achieve impressive results across multiple datasets. However, the security evaluation of such detectors against advanced adversarial attacks is still lacking.

Some prior works~\cite{xie2022evading,huang2020fakepolisher,zijie2023blackbox} focus on the adversarial attack against GAN-generated image detectors. Xie et al~\cite{xie2022evading} proposed a deep dithering approach to effectively eliminate the generative artifacts on various GAN based generated images. In~\cite{huang2020fakepolisher}, a shallow reconstruction method with a learned linear dictionary is proposed to reduce the artifacts incurred during GAN-based fake image synthesis. Lou et al ~\cite{zijie2023blackbox} design a black-box attack through training an anti-forensic encoder-decoder network with contrastive loss. The popular gradient-based adversarial example methods including FGSM~\cite{goodfellow2015explaining}, IFGSM ~\cite{kurakin2018adversarial} and PGD~\cite{madry2018towards} are also used to attack both GAN and DM types of AIGI detectors~\cite{diao2024vulnerabilities,mavali2024fake,de2024exploring}. Diao et al~\cite{diao2024vulnerabilities} demonstrated the vulnerability of AIGI detectors under white- and black-box settings by attacking Convolutional Neural Network (CNN)~\cite{lecun1998gradient} models. Mavali et al~\cite{mavali2024fake} showed that the state-of-the-art AIGI detectors may be defeated even with post-processing in black-box scenario. Rosa et al~\cite{de2024exploring} explored the adversarial robustness of focused on CLIP-based~\cite{radford2021learning} and CNN-based detectors, between which the transferability of attacks is found to be limited. Such prior works show that most AIGI detectors are vulnerable to white-box adversarial attacks. However, the attacks exhibit limited transferability across different detector 

We note that intermediate-level attacks (ILA)~\cite{naseer2018task,wang2021feature,zhang2022improving,li2023improving} is a new adversarial example paradigm with high transferability. Instead of final prediction, the intermediate feature representation which is generally more transferable across architectures is operated. Given advantages of ILA, we creatively introduce it to AIGI anti-forensics. However, existing ILA methods operate solely in spatial domain, ignoring the complementary information in the frequency domain, which has been actually exploited in some prior AIGI detectors~\cite{durall2020watch,zhang2024leveraging}. 

To further improve the transferability, we specifically propose a novel \emph{Dual-domain Feature Importance Attack} (DuFIA) method, which extends the ILA paradigm by jointly leveraging spatial and spectral feature representations. Two complementary feature importance maps are generated from spatial and frequency-domain perturbations, respectively. Such importance maps are fused and integrated into the gradient backward pass, so that the semantically critical and transferable features are selectively emphasized. Unlike the existing ILA methods with static feature weighting, we creatively design a domain-aware mechanism to enhance the cross-model transferability. Extensive experiments demonstrate that DuFIA consistently outperforms existing traditional and ILA attacks in  black-box scenario.

The rest of this letter is organized as follows. The proposed DuFIA scheme is described in Section II, followed by extensive experiments and discussions in Section III. We draw the conclusion in Section IV.

\section{Proposed DuFIA Scheme}

\begin{figure*}[!t]
	\centering
	\includegraphics[width=\textwidth]{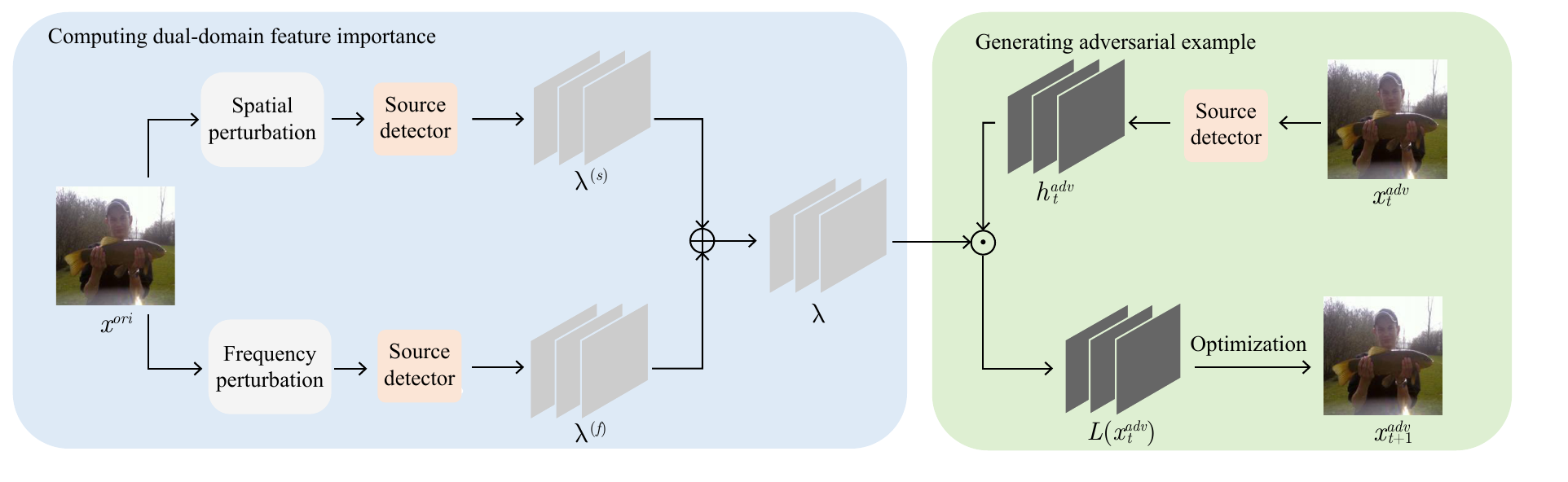}
	\vspace{-0.8cm}
	\caption{Detailed illustration of proposed DuFIA. A original image, after undergoing spatial and frequency domain perturbations, is fed into a source detector, and the dual feature importance at an intermediate layer is obtained via backpropagation. The feature importance is then multiplied element-wise with the same intermediate layer feature map of the adversarial example, producing a loss that guides the generation of the adversarial sample in the next iteration. \(\odot\), \(\oplus\) are the operations of element-wise product and addition, respectively.
	}
	\vspace{-0.2cm}
	\label{fig:Network}
	
\end{figure*}
\subsection{Revisiting Feature Importance in MI-FGSM}
Assume an AIGI detection model \(f_\theta \!: \! x^{ori} \mapsto y \), where \(x^{ori}\) is the original image
and \(y\) is the label of \(x^{ori}\). \(\theta\) denotes the parameters of the model. The goal of adversarial attacks is to yield a perturbed image \(x^{adv}\), which is visually similar to the original image \(x^{ori}\) but makes \( f_\theta(x^{adv}) \ne y \). That can be formulated as a constrained optimization problem
\begin{equation}
	\arg\max_{x^{adv}} J(x^{adv}, y)\,\,\, \text{s.t.}\|x^{adv} - x^{ori}\|_p \leq \epsilon,
\end{equation}
where \(\|\cdot\|_p\) denotes the \(L_p\)-norm measuring the perturbation strength (\(p=\infty\) in this work). \(\epsilon\) is the maximum allowable distortion, and \(J(\cdot,\cdot)\) is the cross entropy loss in AIGI detectors. However, direct optimization of Eq.~(1) requires explicit access to the parameters of \( f_\theta \), which is infeasible in black-box attack scenarios. A feasible alternative is to optimize a surrogate model \( f_\phi \) (the source model), where the parameters are accessible. This approach generates transferable adversarial examples to attack the target model \( f_\theta \).

To solve Eq.(1) in black-box attack scenarios, we select a representative FGSM-inspired boosting method, i.e., MI-FGSM~\cite{dong2018boosting} as our baseline framework. Formally, the adversarial example update at iteration $t$ is generated by
\begin{equation}
	x_{t+1}^{adv} = x_t^{adv} + \alpha \cdot \mathrm{sign}(g_{t+1}),
\end{equation}
\begin{equation}
	g_{t+1} = \mu \cdot g_t + \nabla_{x_t^{adv}} J(x_t^{adv}, y),
\end{equation}
where $g_{t}$ denotes the adversarial image gradient calculated in the $t$-th iteration. The momentum $\mu$ is empirically set as 1.0. $\alpha$ is the step size and $\operatorname{sign}(\cdot)$ denotes the symbolic function.

Different models extract unique features to better adapt to their specific data domains, resulting in model-specific feature representations. When the characteristic is ignored, adversarial attacks often distort features indiscriminately in the source model, leading to local optima specific to that model. Avoiding such local optima is crucial for improving transferability. Adversarial example generation should be guided by key features that are independent of the source model. Feature importance refers to the significance of features that influence the final decision but are independent of the source model.

Inspired by the prior ILA techniques~\cite{wang2021feature,li2023improving} , we compute the intermediate feature importance of AIGI detection networks. That is,
\vspace{-0.2cm}
\begin{equation}
	\nabla_{x_t^{\mathrm{adv}}} J(x_t^{\mathrm{adv}}, y) =
	\sum_{i=0}^{m-1} \lambda_t^{i} \cdot 
	\frac{\partial (h_t^{\mathrm{adv}})_i}{\partial x_t^{\mathrm{adv}}}
	\vspace{-0.2cm}
\end{equation}
with
\begin{equation}
	\lambda_t^{i} = \frac{\partial J(x_t^{adv}, y)}{\partial (h_t^{adv})_i}.
\end{equation}
where $m$ is the number of the activation units in the feature maps. $h_t^{adv}$ represents the mid-layer feature maps of adversarial images at the $t$-th iteration. $\lambda_t^{i}$ denotes the intermediate feature importance weights. During the iterative processing, feature importance $\lambda_t^{i}$ tends to become increasingly dependent on the source model. To mitigate this, a simple yet effective method is proposed in~\cite{wang2021feature}. \(\lambda_t^i\) is set as a constant \(\lambda^i\), which is determined by the original image \(x^{ori}\) rather than the iterative adversarial examples \(x_t^{adv}\) as in MI-FGSM: $\lambda^{i} = \partial J(x^{ori}, y)/\partial (h^{ori})_i$, where $h^{ori}$ denotes the mid-layer feature maps of original images.
\vspace{-0.5cm}
\subsection{Joint Spatial-Frequency Feature Importance}
As illustrated in Fg.~\ref{fig:Network}, we propose a dual-branch feature perturbation strategy to compute the feature importance from both spatial and frequency domains.
\subsubsection{Spatial Perturbation Branch}\hspace{0.5em}
The feature importance $\lambda$ derived from the raw gradients still carries a significant amount of model-specific information. To mitigate this, we employ integrated gradients (IG)~\cite{sundararajan2017axiomatic}. The method aggregates the gradients of interpolated versions of the original image input. The interpolations generate diverse inputs while preserving the spatial structure and overall texture of the image.~\cite{sundararajan2017axiomatic} has shown that, during gradient computation, features related to the decision-making are robust to such transformations, while model-specific features are more susceptible. As a result, the integrated gradients emphasize critical features while suppressing or neutralizing model-specific ones. The spatial-domain feature importance $\lambda^{(s)}$ is once for all computed as 
\begin{equation}
	\lambda^{(s)} = \frac{1}{N} \sum_{i=1}^{N} 
	\frac{\partial J(\frac{i}{N}\, x^{ori} , y)}{\partial h},
\end{equation}
where $h$ denotes the mid-layer feature maps extracted from the interpolated image . \(N\) is the number of integrated steps.
\begin{table*}[t]
	\centering
	\caption{Accuracy of different methods using UnivFD as white-box source and other detectors as black-box targets. The best is \textcolor{red}{\textbf{red}}.}
	\vspace{-0.2cm}
	\renewcommand{\arraystretch}{1.2}
	\setlength{\tabcolsep}{3pt}
	\resizebox{\textwidth}{!}{%
		\begin{tabular}{c|l|cccccc|c|cc|cc|c|ccc|ccc|c|c} 
			\toprule[1.5pt]
			\multirow{2}{*}{Target detectors} & \multirow{2}{*}{Attack methods} 
			& \multicolumn{6}{c|}{GAN-based}  
			& \multirow{2}{*}{Deepfakes} 
			& \multicolumn{2}{c|}{Low-Level} 
			& \multicolumn{2}{c|}{Perceptual loss} 
			& \multirow{2}{*}{Guided}  
			& \multicolumn{3}{c|}{LDM} 
			& \multicolumn{3}{c|}{Glide} 
			& \multirow{2}{*}{DALLE-E} 
			& Total \\ 
			\cline{3-8} \cline{10-11} \cline{12-13} \cline{15-17} \cline{18-20} \cline{22-22}
			& & ProGAN & CycleGAN & BigGAN & StyleGAN & GauGAN & StarGAN 
			& & SITD & SAN & CRN & IMLE & & 200s & 200+CFG & 100s & 100 27 & 50 27 & 100 10 & & Avg. Acc \\ 
			\midrule
			\multirow{7}{*}{UnivFD~\cite{ojha2023towards}}
			& Unattacked & 1.000 & 0.985 & 0.945 & 0.820 & 0.995 & 0.970 & 0.666 &0.630 & 0.575 & 0.595 & 0.720 & 0.700 & 0.942 & 0.738 & 0.944 & 0.791 & 0.799 & 0.781 & 0.867 & 0.802 \\
			& FGSM~\cite{goodfellow2015explaining} & 0.711 & 0.422 & 0.446 & 0.401 & 0.623 & 0.118 & 0.271 &0.322 & 0.317 & 0.290 & 0.319 & 0.504 & 0.511 & 0.423 & 0.516 & 0.528 & 0.538 & 0.530 & 0.460 & 0.418 \\
			& PGD~\cite{madry2018towards} & 0.412 & 0.184 & 0.258 & 0.236 & 0.360 & 0.730 & 0.026 &0.117 & 0.381 & 0.301 & 0.320 & 0.430 & 0.475 & 0.431 & 0.473 & 0.434 & 0.438 & 0.438 & 0.460 & 0.386 \\
			& C\&W~\cite{carlini2017towards} & 0.967 & 0.917 & 0.784 & 0.721 & 0.954 & 0.831 & 0.605 &0.591 & 0.498 & 0.575 & 0.627 & 0.627 & 0.837 & 0.628 & 0.844 & 0.698 & 0.703 & 0.698 & 0.724 & 0.729 \\
			& MIFGSM~\cite{dong2018boosting} & 0.213 & 0.054 & 0.074 & 0.055 & 0.142 & 0.001 & 0.037 &0.112 & 0.144 & 0.075 & 0.140 & 0.317 & 0.338 & 0.312 & 0.334 & 0.325 & 0.318 & 0.327 & 0.333 & \textcolor{red}{\textbf{0.210}} \\
			& FIA~\cite{wang2021feature} & 0.330 & 0.154 & 0.156 & 0.154 & 0.273 & 0.045 & 0.082 &0.121 & 0.267 & 0.215 & 0.272 & 0.375 & 0.394 & 0.382 & 0.385 & 0.375 & 0.381 & 0.393 & 0.395 & 0.280 \\
			& DuFIA (Ours) & 0.258 & 0.099 & 0.085 & 0.070 & 0.172 & 0.013 & 0.052 &0.125 & 0.135 & 0.054 & 0.085 & 0.368 & 0.378 & 0.365 & 0.371 & 0.362 & 0.367 & 0.366 & 0.363 & 0.220 \\
			\hline
			\multirow{7}{*}{Corvi~\cite{corvi2023detection}} 
			& Unattacked & 0.511 & 0.463 & 0.519 & 0.598 & 0.506 & 0.457 & 0.566 &0.781  & 0.801 & 0.500 & 0.500 & 0.514 & 0.993 & 0.993 & 0.580 & 0.622 & 0.591 & 0.514 & 0.894 & 0.592 \\
			& FGSM~\cite{goodfellow2015explaining}      & 0.504 & 0.495 & 0.502 & 0.503 & 0.501 & 0.482 & 0.495 &0.671 & 0.498 & 0.500 & 0.500 & 0.487 & 0.964 & 0.947 & 0.963 & 0.490 & 0.490 & 0.491 & 0.585 & 0.565  
			\\
			& PGD~\cite{madry2018towards}    & 0.503 & 0.480 & 0.500 & 0.503 & 0.503 & 0.466 & 0.434 &0.517 & 0.505 & 0.500 & 0.500 & 0.377 & 0.978 & 0.979 & 0.572 & 0.5155 & 0.498 &0.505 & 0.563 & 0.534
			\\
			& C\&W~\cite{carlini2017towards}   & 0.512 & 0.461 & 0.501 & 0.518 & 0.501 & 0.454 & 0.539 &0.497 & 0.546 & 0.500 & 0.500 & 0.4645 & 0.989 & 0.989 & 0.989 & 0.531 & 0.542 & 0.550 & 0.782 & 0.576        
			\\
			& MIFGSM~\cite{dong2018boosting}     & 0.502 & 0.484 & 0.499 & 0.501 & 0.501 & 0.448 & 0.371 &0.469  & 0.507 & 0.500 & 0.500 & 0.487 & 0.957 & 0.942 & 0.489 & 0.495 & 0.490 & 0.487 & 0.567 & 0.514 \\
			& FIA~\cite{wang2021feature}        & 0.501 & 0.490 & 0.499 & 0.500 & 0.500 & 0.470 & 0.441 &0.442  & 0.527 & 0.500 & 0.500 & 0.488 & 0.944 & 0.915 & 0.490 & 0.490 & 0.490 & 0.488 & 0.560 & 0.509 \\
			& DuFIA (Ours)      & 0.510 & 0.461 & 0.505 & 0.508 & 0.516 & 0.221 & 0.251 &0.415  & 0.484 & 0.500 & 0.500 & 0.487 & 0.936 & 0.901 & 0.486 & 0.490 & 0.489 & 0.487 & 0.556 & \textcolor{red}{\textbf{0.495}} \\
			\hline
			\multirow{7}{*}{Cozz~\cite{cozzolino2024raising}} 
			& Unattacked & 0.726 & 0.873 & 0.741 & 0.705 & 0.839 & 0.550 & 0.510 &0.823 & 0.776 & 0.531 & 0.537 & 0.439 & 0.691 & 0.681 & 0.677 & 0.681 & 0.667 & 0.678 & 0.653 & 0.677 \\
			& FGSM~\cite{goodfellow2015explaining} & 0.581 & 0.495 & 0.527 & 0.497 & 0.549 & 0.501 & 0.467 &0.484 & 0.461 & 0.524 & 0.550 & 0.475 & 0.638 & 0.630 & 0.639 & 0.587 & 0.594 & 0.592 & 0.609 & 0.574     
			\\
			& PGD~\cite{madry2018towards}  & 0.575 & 0.480 & 0.500 & 0.497 & 0.561 & 0.482 & 0.461 &0.512 & 0.468 & 0.538 & 0.538 & 0.374 & 0.605 & 0.585 & 0.599 & 0.508 & 0.479 & 0.499 & 0.617 & 0.520       
			\\
			& C\&W~\cite{carlini2017towards}   & 0.623 & 0.461 & 0.500 & 0.518 & 0.767 & 0.454 & 0.513 &0.498 & 0.530 & 0.526 & 0.526 & 0.465 & 0.605 & 0.602 & 0.607 & 0.596 & 0.586 & 0.592 & 0.781 & 0.571        
			\\
			& MIFGSM~\cite{dong2018boosting}     & 0.531 & 0.511 & 0.459 & 0.448 & 0.461 & 0.490 & 0.487 &0.473 & 0.463 & 0.475 & 0.496 & 0.375 & 0.584 & 0.584 & 0.549 & 0.552 & 0.546 & 0.554 & 0.557 & 0.506 \\
			& FIA~\cite{wang2021feature}         & 0.513 & 0.519 & 0.452 & 0.431 & 0.458 & 0.481 & 0.478 &0.467 & 0.482 & 0.453 & 0.466 & 0.433 & 0.599 & 0.593 & 0.556 & 0.556 & 0.548 & 0.551 & 0.556 & 0.510 \\
			& DuFIA (Ours)       & 0.524 & 0.493 & 0.474 & 0.435 & 0.458 & 0.484 & 0.489 &0.427 & 0.452 & 0.390 & 0.409 & 0.424 & 0.593 & 0.582 & 0.554 & 0.553 & 0.553 & 0.552 & 0.560 & \textcolor{red}{\textbf{0.505}} \\
			\hline
			\multirow{7}{*}{DRCT~\cite{chen2024drct}} 
			& Unattacked & 0.801 & 0.945 & 0.827 & 0.756 & 0.784 & 0.630 & 0.573 &0.810  & 0.788 & 0.439 & 0.505 & 0.785 & 0.866 & 0.816 & 0.868 & 0.84 & 0.833 & 0.848 & 0.86 & 0.767 \\
			& FGSM~\cite{goodfellow2015explaining} & 0.610 & 0.4940 & 0.501 & 0.438 & 0.576 & 0.414 & 0.497 &0.497  & 0.477 & 0.561 & 0.661 & 0.373 & 0.478 & 0.461 & 0.486 & 0.498 & 0.498 & 0.506 & 0.477 & 0.489    
			\\
			& PGD~\cite{madry2018towards}  & 0.530   & 0.436 & 0.426 & 0.332 & 0.519 & 0.330 & 0.318 &0.524  & 0.479 & 0.375 & 0.479 & 0.262 & 0.451 & 0.410  & 0.463 & 0.347 & 0.347 & 0.366 & 0.405 & 0.411        
			\\
			& C\&W~\cite{carlini2017towards} & 0.693 & 0.734 & 0.733 & 0.500 & 0.856 & 0.528 & 0.517 &0.514  & 0.605 & 0.685 & 0.796 & 0.488 & 0.677 & 0.600 & 0.685 & 0.670 & 0.657 & 0.681 & 0.648 & 0.684    
			\\
			& MIFGSM~\cite{dong2018boosting}     & 0.486 & 0.414 & 0.383 & 0.341 & 0.424 & 0.371 & 0.422 &0.429  & 0.488 & 0.35 & 0.419 & 0.665 & 0.649 & 0.601 & 0.678 & 0.6 & 0.672 & 0.613 & 0.601 & 0.517 \\
			& FIA~\cite{wang2021feature}        & 0.292 & 0.308 & 0.292 & 0.261 & 0.355 & 0.288 & 0.411 &0.415  & 0.397 & 0.158 & 0.238 & 0.49 & 0.56 & 0.511 & 0.559 & 0.549 & 0.529 & 0.574 & 0.534 & 0.404 \\
			& DuFIA (Ours)       & 0.213 & 0.282 & 0.262 & 0.253 & 0.283 & 0.215 & 0.264 &0.398  & 0.477 & 0.204 & 0.287 & 0.436 & 0.533 & 0.51 & 0.533 & 0.483 & 0.469 & 0.498 & 0.537 & \textcolor{red}{\textbf{0.376}} \\
			\hline
			\multirow{7}{*}{Rajan~\cite{rajan2024aligned}} 
			& Unattacked & 0.523 & 0.493 & 0.512 & 0.533 & 0.507 & 0.518 & 0.509 &0.612  & 0.564 & 0.501 & 0.500 & 0.516 & 0.995 & 0.995 & 0.995 & 0.656 & 0.695 & 0.679 & 0.783 & 0.599 \\
			& FGSM~\cite{goodfellow2015explaining} & 0.503 & 0.493 & 0.502 & 0.506 & 0.501 & 0.497 & 0.481 &0.513  & 0.500 & 0.500 & 0.500 & 0.495 & 0.989 & 0.984 & 0.988 & 0.511 & 0.520 & 0.508 & 0.600 & 0.557 \\
			& PGD~\cite{madry2018towards}  & 0.502 & 0.495 & 0.500 & 0.501 & 0.500 & 0.500 & 0.499 &0.521  & 0.507 & 0.500 & 0.500 & 0.494 & 0.992 & 0.985 & 0.992 & 0.496 & 0.495 & 0.497 & 0.578 & 0.556 \\
			& C\&W~\cite{carlini2017towards} & 0.511 & 0.492 & 0.504 & 0.509 & 0.501 & 0.506 & 0.484 &0.509  & 0.498 & 0.500 & 0.500 & 0.502 & 0.995 & 0.995 & 0.995 & 0.561 & 0.586 & 0.572 & 0.704 & 0.575 \\   
			& MIFGSM~\cite{dong2018boosting}     & 0.502 & 0.495 & 0.498 & 0.501 & 0.500 & 0.499 & 0.495 &0.500  & 0.502 & 0.500 & 0.500 & 0.492 & 0.979 & 0.971 & 0.981 & 0.495 & 0.498 & 0.494 & 0.561 & 0.551 \\
			& FIA~\cite{wang2021feature}        & 0.501 & 0.497 & 0.500 & 0.500 & 0.500 & 0.500 & 0.500 &0.498  & 0.498 & 0.500 & 0.500 & 0.496 & 0.978 & 0.943 & 0.978 & 0.495 & 0.496 & 0.498 & 0.548 & 0.549 \\
			& DuFIA (Ours)       & 0.511 & 0.493 & 0.504 & 0.501 & 0.507 & 0.498 & 0.487 &0.493  & 0.502 & 0.500 & 0.500 & 0.496 & 0.969 & 0.931 & 0.962 & 0.495 & 0.499 & 0.498 & 0.554 & \textcolor{red}{\textbf{0.547}} \\
			\hline
			\multirow{7}{*}{RINE~\cite{koutlis2024leveraging}} 
			& Unattacked &1.000  &0.993  &0.996  &0.889  &0.998  &0.995  &0.806  &0.906  &0.683  &0.892  &0.906  &0.761  &0.983  &0.882  &0.986  &0.889  &0.926  &0.907  &0.950  &0.861  \\
			& FGSM~\cite{goodfellow2015explaining} & 0.855 & 0.659 & 0.604 & 0.555 & 0.804 & 0.587 & 0.527 &0.521  & 0.457 & 0.373 & 0.445 & 0.565 & 0.563 & 0.495 & 0.572 & 0.539 & 0.544 & 0.541 & 0.526 & 0.562 \\
			& PGD~\cite{madry2018towards}  & 0.663 & 0.485 & 0.459 & 0.369 & 0.652 & 0.221 & 0.108 &0.501  & 0.466 & 0.201 & 0.361 & 0.471 & 0.534 & 0.478 & 0.541 & 0.489 & 0.489 & 0.484 & 0.548 & 0.422 \\
			& C\&W~\cite{carlini2017towards} & 0.991 & 0.979 & 0.965 & 0.878 & 0.987 & 0.970 & 0.796 &0.573  & 0.523 & 0.683 & 0.684 & 0.699 & 0.941 & 0.817 & 0.945 & 0.861 & 0.908 & 0.880 & 0.912 & 0.839 \\
			& MIFGSM~\cite{dong2018boosting}     &0.453  &0.299  &0.281  &0.251  &0.454  &0.202  &0.122  &0.419  &0.380  &0.446  &0.466  &0.492  &0.444  &0.505  &0.455  &0.454  &0.460  &0.484  &0.388  &0.386  \\
			& FIA~\cite{wang2021feature}        &0.480  &0.361  &0.253  &0.214  &0.426  &0.265  &0.172  &0.352  &0.320  &0.399  &0.473  &0.528  &0.455  &0.550  &0.474  &0.473  &0.476  &0.507  &0.399  &0.399  \\
			& DuFIA (Ours)  &0.412  &0.235  &0.217  &0.223  &0.371  &0.170  &0.194  &0.317  &0.318  &0.374  &0.520  &0.525  &0.423  &0.495  &0.494  &0.490  &0.511  &0.490  &0.384  &\textcolor{red}{\textbf{0.377}}  \\
			\bottomrule[1.5pt]
	\end{tabular}}
	\label{tab:attack_accuracy}
	\vspace{-0.4cm}
\end{table*}
\begin{figure}[!t]
	\centering
	\includegraphics[width=\linewidth]{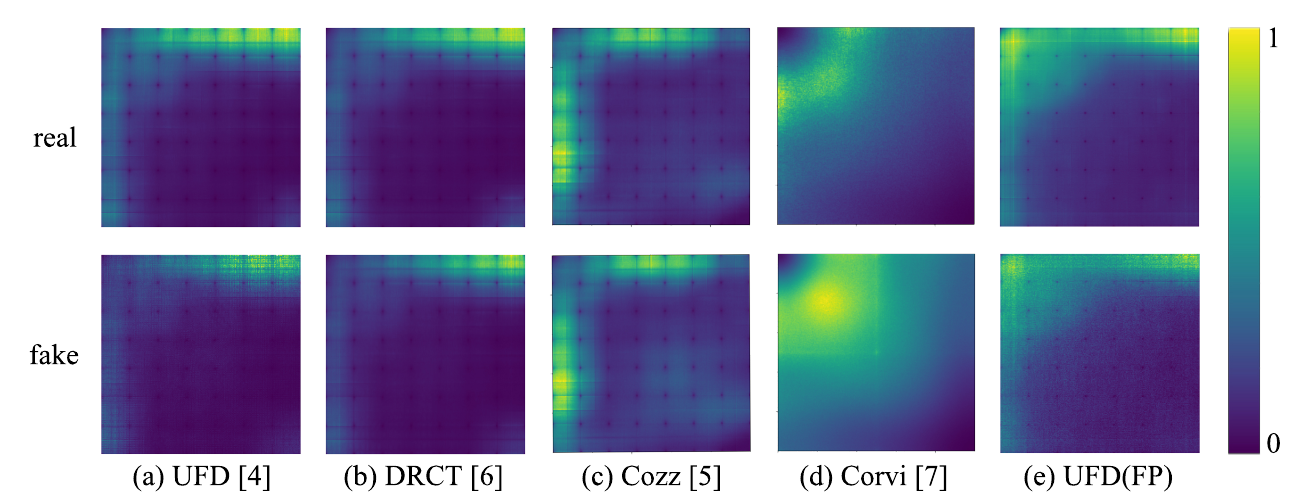}
	\caption{Spectrum saliency maps (averaged over all images from the CycleGAN dataset~\cite{ojha2023towards}) for different AIGI detectors. (a-d) the results of the raw images for four different  detectors. (e) the result of the images with frequency perturbation (FP) for UFD~\cite{ojha2023towards}.}
	\label{fig:ssm}
	\vspace{-0.6cm}
\end{figure} 
\subsubsection{Frequency Perturbation Branch} \hspace{0.1em}
Previous studies~\cite{Frank2020, Jeong2022} have shown that real and synthetic images exhibit distinct differences in the frequency domain. Additionally, some prior AIGI detectors~\cite{durall2020watch}~\cite{ zhang2024leveraging} have also leveraged frequency domain information. To this end, we calculate the feature importance in the frequency domain. Similar to spatial perturbation, we apply a transformation to the original image in the frequency domain~\cite{long2022frequency}:
\begin{equation}
	x^{(f)} = \operatorname{IDCT} \left( M \odot \operatorname{DCT}(x^{ori} + \boldsymbol{\xi}) \right),
\end{equation}
where the discrete cosine transform (DCT) and the inverse discrete cosine transform (IDCT) are applied. \(M \sim \mathcal{U}(1 - p, 1 + p)\) is a random frequency mask drawn from a uniform distribution \(\mathcal{U}(\cdot)\). \(p\) controls the perturbation range of each frequency coefficient. \(\odot\) denotes the operation of element-wise product and \(\boldsymbol{\xi} \sim \mathcal{N}(0, \sigma^2)\) is Gaussian noise. The frequency-domain feature importance is then estimated as
\begin{equation}
	\lambda^{(f)} =  \frac{\partial J(x^{(f)}, y)}{\partial h}.
\end{equation}

We use the spectrum saliency map~\cite{long2022frequency} computed as \(\partial J\big(\text{IDCT}(\text{DCT}(x)), y\big) / \partial \text{DCT}(x)\) to highlight frequency components. In Fig.~\ref{fig:ssm}, low-frequency components are mainly concentrated in the upper left corner, while high-frequency components are distributed in the lower right corner. Fig.~\ref{fig:ssm} illustrates two key phenomena: (1) In the frequency domain, there are significant differences between real and synthetic images. Adding perturbations can reduce the gap between the differences, making it more difficult for detectors to distinguish. (2) Different detectors tend to exploit different frequency features for classification. Random frequency perturbations help reduce the discrepancy among detectors.
\subsubsection{Joint Feature Importance} \hspace{0.1em}
Finally, joint feature importance is computed by
\begin{equation}
	\lambda = \frac{ \lambda^{(s)} + \lambda^{(f)} }{2}.
	\vspace{-0.2cm}
\end{equation}
\subsection{Generating Adversarial Example via DuFIA Objective}

As is illustrated in Fig.~\ref{fig:Network}, instead of the cross-entropy loss \(J(\cdot,\cdot)\) as MI-FGSM, DuFIA essentially works by maximizing a novel mid-layer feature attack loss:
\begin{equation}
	L(x_t^{adv},y) = \sum (\lambda \odot h_t^{adv}),
\end{equation}
The optimization objective of DuFIA can be defined as
\begin{equation}
	\arg\max_{x^{adv}} L(x^{adv}, y)\,\,\, \text{s.t.}\|x^{adv} - x^{ori}\|_p \leq \epsilon.
\end{equation}

\section{Experiments}
\label{sec:experiments}

\subsection{Experimental Settings}

\subsubsection{Datasets}
{
\hspace{0.01em}
We evaluate the performance of attack schemes on a large scale of public AI-generated image dataset, which has been used as benchmark in prior works~\cite{ojha2023towards,koutlis2024leveraging}. Specifically, the synthetic image are created by the generated models including ProGAN, StyleGAN, StyleGAN2, BigGAN, CycleGAN~, StarGAN, GauGAN, DeepFake, SITD, SAN, CRN, IMLE, Guided Diffusion, LDM, Glide, and DALL-E. Most of the images have a spatial resolution of 256 × 256 pixels.
}

\subsubsection{Target Detectors}
{
\hspace{0.01em}
To evaluate detection performance and adversarial robustness, we select six representative state-of-the-art AI-generated image detectors spanning diverse architectures and training approaches. That is, UnivFD~\cite{ojha2023towards}, Corvi~\cite{corvi2023detection}, Cozz~\cite{cozzolino2024raising}, DRCT~\cite{chen2024drct}, Rajan~\cite{rajan2024aligned}, and RINE~\cite{koutlis2024leveraging}.
}
\subsubsection{Compared Attacks}
{
\hspace{0.01em}
Six representative adversarial attack methods. FGSM~\cite{goodfellow2015explaining}, PGD~\cite{madry2018towards}, Carlini \& Wagner~\cite{carlini2017towards}, MI-FGSM~\cite{dong2018boosting} and FIA~\cite{wang2021feature} in the test.In pursuit of fairness
and comprehensiveness, we followed previous works~\cite{mavali2024fake,diao2024vulnerabilities} and made efforts to include the attacks they had used. However, since some works did not release their source code, we could only use limited attack methods from their papers.
}

\subsubsection{Evaluation Metrics}
{
\hspace{0.01em}
Following previous works~\cite{ojha2023towards,koutlis2024leveraging}, we evaluate the classification accuracy of of AIGI detectors based on a fixed threshold 0.5. PSNR, SSIM and LPIPS~\cite{mavali2024fake} are adopted to measure the  visual quality. 
}

\subsubsection{Implementation details}
{
\hspace{0.01em}
 To ensure imperceptible perturbation and fair comparison with counterparts, we set the maximum perturbation \(\epsilon = 8/255\), the step size \(\alpha = 0.8/255\), and the number of iterations 10.All attacks use the $\ell_{\infty}$ norm. For C\&W, we set the number of iterations  $500$ and the learning rate $0.005$. MIFGSM, FIA and DuFIA are enhanced with momentum factor \(\mu = 1.0\). We choose the 5th transformer block as the intermediate layer for ViT-based models.
}

\begin{table}[!t]
	\caption{Perceptual quality metrics of different attacks.}
	\vspace{-0.2cm}
	\renewcommand{\arraystretch}{1.2} 
	\centering
	\setlength{\tabcolsep}{1pt}      
	\begin{adjustbox}{width=\linewidth}
		\begin{tabular}{l|cccccc}
			\toprule[1.5pt]
			Attack & FGSM~\cite{goodfellow2015explaining} & PGD~\cite{madry2018towards} & C\&W~\cite{carlini2017towards} & MIFGSM~\cite{dong2018boosting} & FIA~\cite{wang2021feature} & DuFIA (Ours) \\ 
			\midrule
			PSNR (dB) $\uparrow$ & 33.28 & 32.64 & 33.31 & 33.16 & 32.48 & \textcolor{red}{\textbf{33.42}} \\
			SSIM $\uparrow$     & 0.870 & 0.857 & 0.874 & 0.873 & 0.854 & \textcolor{red}{\textbf{0.881}} \\
			LPIPS $\downarrow$    & 0.072 & 0.087 & 0.075 & 0.067 & 0.089 & \textcolor{red}{\textbf{0.062}} \\
			\bottomrule[1.5pt]
		\end{tabular}
	\end{adjustbox}
	\label{tab:quality_metrics}
\end{table}

\vspace{-0.4cm}
\subsection{Comparison with State-of-the-Art Methods}
Table~\ref{tab:attack_accuracy} presents the performance of different attack methods using UnivFD as the source detector and all other AIGI detectors as targets. Each row corresponds to the adversarial examples crafted on UnivFD, while each column shows the accuracy when transferred to a specific target detector. DuFIA consistently results in the lowest average accuracy, indicating its superior transferability. Table~\ref{tab:quality_metrics} compares the average perceptual quality metrics of all adversarial samples yielded by different attacks.The results show that DuFIA achieves the highest PSNR and SSIM values and the lowest LPIPS score among all attack methods. It implies the best visual quality for the attacked images yielded by our scheme.
\vspace{-0.5cm}
\subsection{Robustness Evaluation}
We further evaluate the robustness of attack methods against common post image degradations, which frequently occur in real-world application, such as social media platform. Specifically,three types of degradations are applied to adversarial examples: JPEG compression with quality factor Q=90/60/30, Gaussian blur with standard deviations (0.02/0.08/0.64), and additive Gaussian noise with intensities of 1/255, 8/255 and 64/255.Table~\ref{tab:acc_degradation} shows the average of RINE detector on the post-processed adversarial example images. As shown, DuFIA consistently achieves the lowest accuracy across most scenarios, indicating better robustness against post degradation.

\begin{table}[t]
	\centering
	\caption{Accuracy of RINE~\cite{koutlis2024leveraging} on the attacked CycleGAN dataset against different post-processing when attacking UnivFD.}
	\vspace{-0.2cm}
	\renewcommand{\arraystretch}{1.2} 
	\setlength{\tabcolsep}{5pt}       
	\begin{adjustbox}{width=\linewidth}
		\begin{tabular}{l|ccc|ccc|ccc}
			\toprule[1.5pt]
			Methods & \multicolumn{3}{c|}{JPEG} & \multicolumn{3}{c|}{Blur} & \multicolumn{3}{c}{Noise} \\ 
			& 90 & 60 & 30 & 0.02 & 0.32 & 0.64 & 1/255 & 8/255 & 64/255 \\ 
			\midrule
			Unattacked & 0.987 & 0.900 & 0.885 & 0.993 & 0.992 & 0.993 & 0.975 & 0.903 & 0.651 \\
			FGSM~\cite{goodfellow2015explaining}       & 0.681 & 0.682 & 0.717 & 0.659 & 0.659 & 0.679 & 0.658 & 0.676 & 0.624 \\
			PGD~\cite{madry2018towards}        & 0.736 & 0.749 & 0.803 & 0.485 & 0.485 & 0.606 & 0.480 & 0.667 & 0.625 \\
			C\&W~\cite{carlini2017towards}     & 0.982 & 0.965 & 0.881 & 0.979 & 0.979 & 0.978 & 0.969 & 0.860 & 0.636 \\
			MIFGSM~\cite{dong2018boosting}     & 0.405 & 0.563 & 0.657 & 0.299 & 0.312 & 0.383 & 0.289 & 0.446 & 0.611 \\
			FIA~\cite{wang2021feature}        & 0.457 & 0.547 & 0.596 & 0.361 & 0.378 & 0.481 & 0.347 & 0.436 & \textcolor{red}{\textbf{0.563}} \\
			DuFIA (Ours)      & \textcolor{red}{\textbf{0.301}} & \textcolor{red}{\textbf{0.445}} & \textcolor{red}{\textbf{0.551}} & 
			\textcolor{red}{\textbf{0.235}} & \textcolor{red}{\textbf{0.248}} & \textcolor{red}{\textbf{0.293}} & 
			\textcolor{red}{\textbf{0.224}} & \textcolor{red}{\textbf{0.328}} & 0.591 \\
			\bottomrule[1.5pt]
		\end{tabular}
	\end{adjustbox}
	\label{tab:acc_degradation}
\end{table}

\begin{table}[!t]
	\centering
	\caption{Ablation study on the CycleGAN dataset. Adversarial samples are from UnivFD~\cite{ojha2023towards} and evaluated on different models by accuracy.}
	\vspace{-0.2cm}
	\renewcommand{\arraystretch}{1.2} 
	\setlength{\tabcolsep}{1pt}       
	\begin{adjustbox}{width=\linewidth}
		\begin{tabular}{lccccccc}
			\toprule[1.5pt]
			Attack & UnivFD~\cite{ojha2023towards} & Corvi~\cite{corvi2023detection} & Cozz~\cite{cozzolino2024raising} & DRCT~\cite{chen2024drct} & Rajan~\cite{rajan2024aligned} & RINE~\cite{koutlis2024leveraging} & Avg \\ 
			\midrule
			None Perturbation              & 0.154 & 0.490 & 0.519 & 0.308 & 0.497 & 0.361 & 0.388 \\
			Spatial Perturbation           & 0.150 & 0.473 & 0.509 & 0.289 & \textcolor{red}{\textbf{0.492}} & 0.254 & 0.361 \\
			Frequency Perturbation         & \textcolor{red}{\textbf{0.085}} & 0.491 & 0.495 & 0.284 & 0.497 & 0.240 & 0.349 \\
			Spatial-Frequency Perturbation & 0.099 & \textcolor{red}{\textbf{0.461}} & \textcolor{red}{\textbf{0.493}} & \textcolor{red}{\textbf{0.282}} & 0.493 & \textcolor{red}{\textbf{0.235}} & \textcolor{red}{\textbf{0.344}} \\
			\bottomrule[1.5pt]
		\end{tabular}
	\end{adjustbox}
	\label{tab:ablation}
\end{table}
\vspace{-0.4cm}
\subsection{Ablation Study}
We conduct an ablation study to investigate the impact of perturbation in different domains. As shown in Table~\ref{tab:ablation}, computing gradients solely in the spatial domain or frequency domain leads to lower transferability. Our proposed joint spatial-frequency strategy adopted in DuFIA achieves the best.

\section{Conclusion}
This work proposes DuFIA, a dual-domain feature importance attack that integrates spatial and frequency perturbations. By adaptively fusing complementary feature gradients, DuFIA significantly improves transferability against diverse AI-generated image detectors. Extensive experiments verify its effectiveness under various AIGI detectors and degradations. In future work, we plan to refine the adaptive fusion mechanism, extend DuFIA to multimodal tasks.
\label{sec:conclusion}

\bibliographystyle{IEEEtran}
\bibliography{ref}

\end{document}